\documentclass[runningheads]{llncs}
\usepackage[T1]{fontenc}
\usepackage{amsmath}
\usepackage{amssymb}
\usepackage{graphicx}
\usepackage{booktabs}
\usepackage{multirow}
\usepackage{siunitx}
\usepackage{subfig}

\begin{document}
\title{Disentangling Model and Human Data Uncertainty in Apparent Facial Age Estimation}
\titlerunning{Disentangling Human Data Uncertainty in Apparent Facial Age Estimation}
\author{Andrei Foitos \and Ivo Pascal de Jong \and Matias Valdenegro-Toro}
\authorrunning{A. Foitos et al.}
\institute{Bernoulli Institute, University of Groningen, Groningen, NL}
\maketitle              %
\begin{abstract}
Estimating the apparent age of individuals from facial images is challenging due to the subjective nature of perception and the inherent variability of the data. We investigate the role of uncertainty estimation, attributing uncertainty jointly to a lack of knowledge (epistemic) or inherent noise/chance (aleatoric). Leveraging the APPA-REAL dataset, we train Bayesian Neural Networks on datasets of varying sizes using three BNN approximations: MC-DropConnect, Flipout, and Deep Ensembles using supervision on human aleatoric uncertainty available in the APPA-REAL dataset. Each model outputs both the predicted apparent age and the amount of aleatoric and epistemic uncertainty. Our results confirm the hypothesis that the inherent aleatoric uncertainty remains stable across dataset sizes, while epistemic uncertainty increases as training data decreases. These findings demonstrate that different sources of uncertainty can be quantified in face age estimation. 
\end{abstract}

\section{Introduction}
\label{sec:introduction}

Artificial intelligence and machine learning systems are becoming increasingly integral to real-world decision-making \cite{mckinsey2024aisurvey}. While the predictive capabilities of modern models are impressive, their adoption in sensitive applications often hinges on more than just accuracy. It is also required to understand the reliability of those predictions. In high-stakes contexts, a wrong or overly confident prediction can have serious consequences. Thus, there is a growing demand for machine learning models that not only make predictions but also quantify their uncertainty \cite{lakshminarayanan2017simple} \cite{gawlikowski2021survey} \cite{ribeiro2016why}.

One application area where prediction uncertainty is particularly relevant is facial age estimation, the task of predicting a person's age based solely on a facial image \cite{fu2010survey}. These models estimate the apparent age of a person, based on visual features of aging, rather than their actual chronological age \cite{escalera2015chalearn}. Apparent age estimation introduces a layer of complexity: apparent age is subjective and often has a large degree of uncertainty. Human annotators may disagree due to individual bias or cultural perception \cite{agustsson2017appa} \cite{escalera2017chalearn}, or consider multiple answers to be valid. 

Machine learning regression models typically only output a single most likely value. In scenarios where there is no single correct answer -- including apparent age estimation -- this is insufficient. Instead, a model should give a most-likely estimate, and estimate the variance of which ages are likely. This is important for applications where age estimation is used to support the enforcement of age-restricted content or purchases.

With Uncertainty Quantification (UQ), the model not only outputs the most likely age but also provides an estimate of uncertainty \cite{he2023survey}. There are two causes why a Machine Learning model (or anything) can be uncertain.

\begin{figure}[t]
    \centering
    \subfloat{
        \centering
        \includegraphics[width=0.49\linewidth]{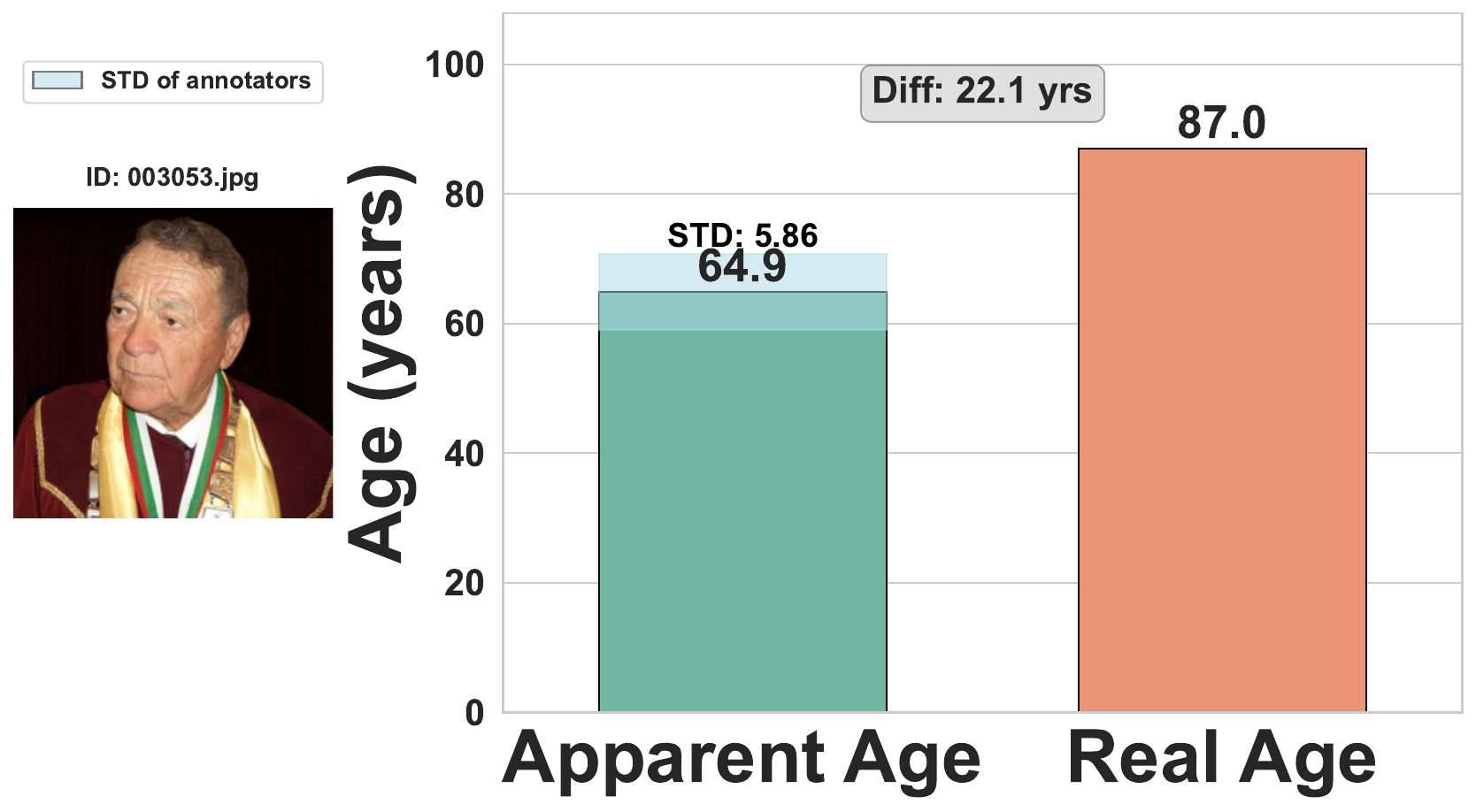}
        \label{fig:old}
    }
    \subfloat{
        \centering
        \includegraphics[width=0.49\linewidth]{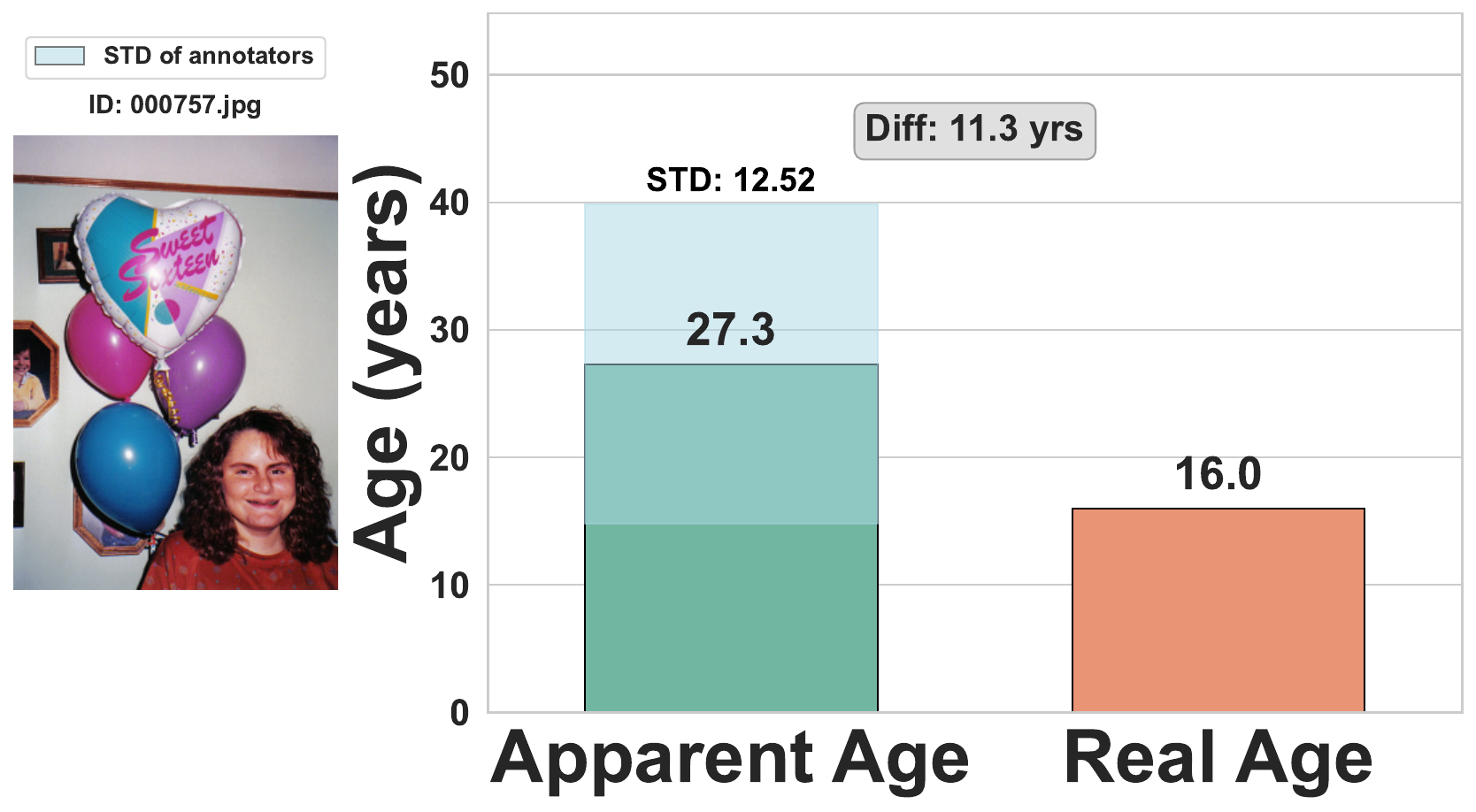}
        \label{fig:young}
    }
    \caption{Comparison of old (left) and young (right) apparent age estimation examples with uncertainty estimation. This paper proposes to learn human aleatoric uncertainty separately from epistemic uncertainty.}
\end{figure}

While methods for disentangling these two types of uncertainty have been around for longer \cite{kendall2017uncertainties} \cite{H_llermeier_2021}, recent work has found that in classification contexts the two types of uncertainty are not successfully separated \cite{mucsanyi2024benchmarking},\cite{wimmer2023quantifying} \cite{dejong2025uncertainty}. Since age estimation is a regression task, it is unknown whether these critiques on disentanglement apply or whether each uncertainty can be successfully isolated. 

\textbf{Contributions}. We address the gap in a standard age prediction model by combining age regression with uncertainty quantification. Knowing only what a model predicts is insufficient; we must also understand how uncertain the model is, and why it is uncertain. And particularly in facial age estimation, there is a high degree of uncertainty, both coming from the model itself and labeling uncertainty, via the concept of apparent age, different humans will label face image with different age values, so a successful facial age estimation model must estimate both model and data uncertainties. This formulation also learns from multiple human annotations (available in the APPA-REAL dataset), incorporating human uncertainty into the trained model.

\begin{figure*}[!t]
    \centering
    \subfloat[Scatterplot]{
        \centering
        \includegraphics[width=0.32\linewidth]{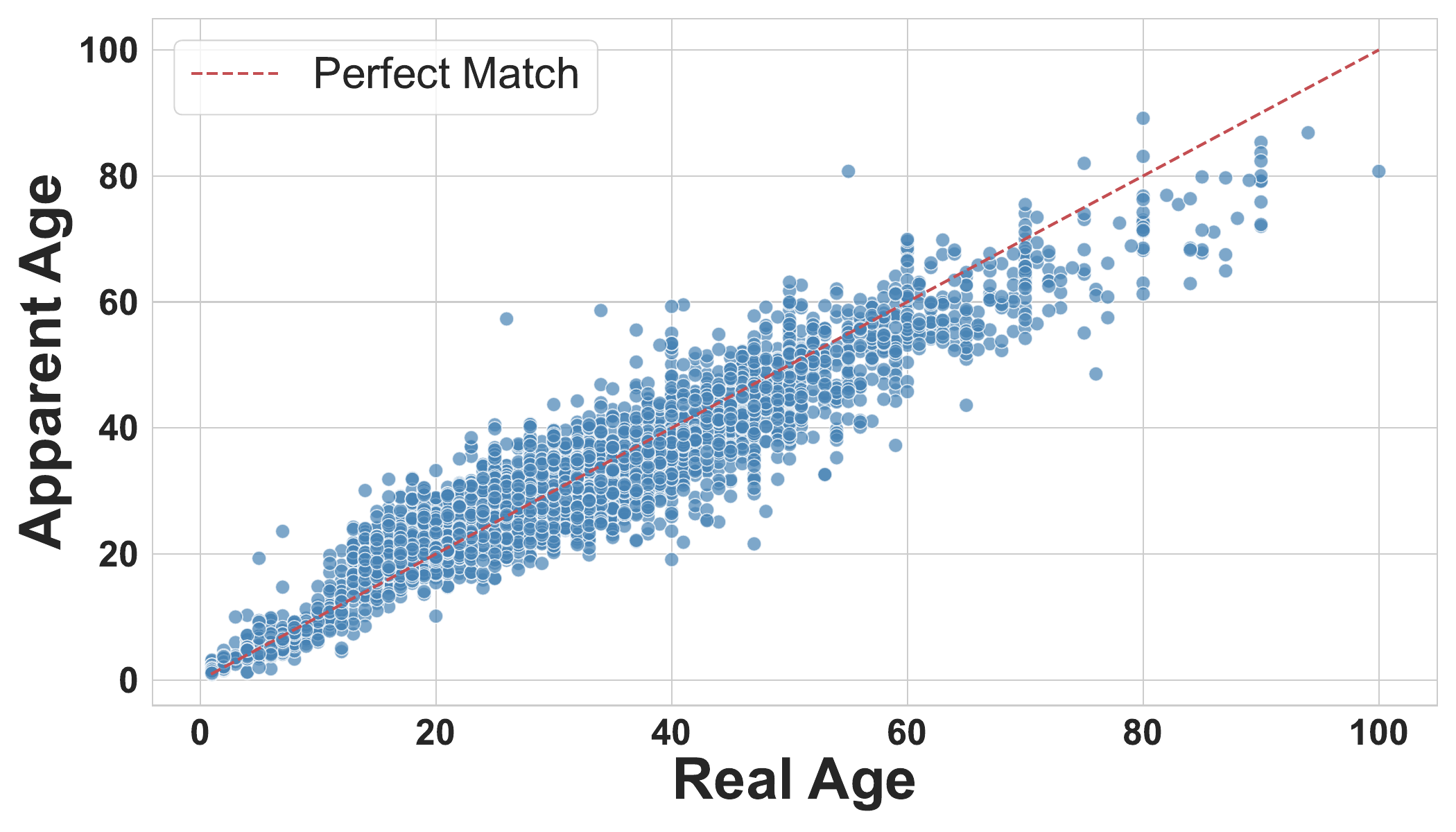}
        \label{fig:realvsapp}
    }
    \subfloat[Age Distribution]{
        \centering
        \includegraphics[width=0.32\linewidth]{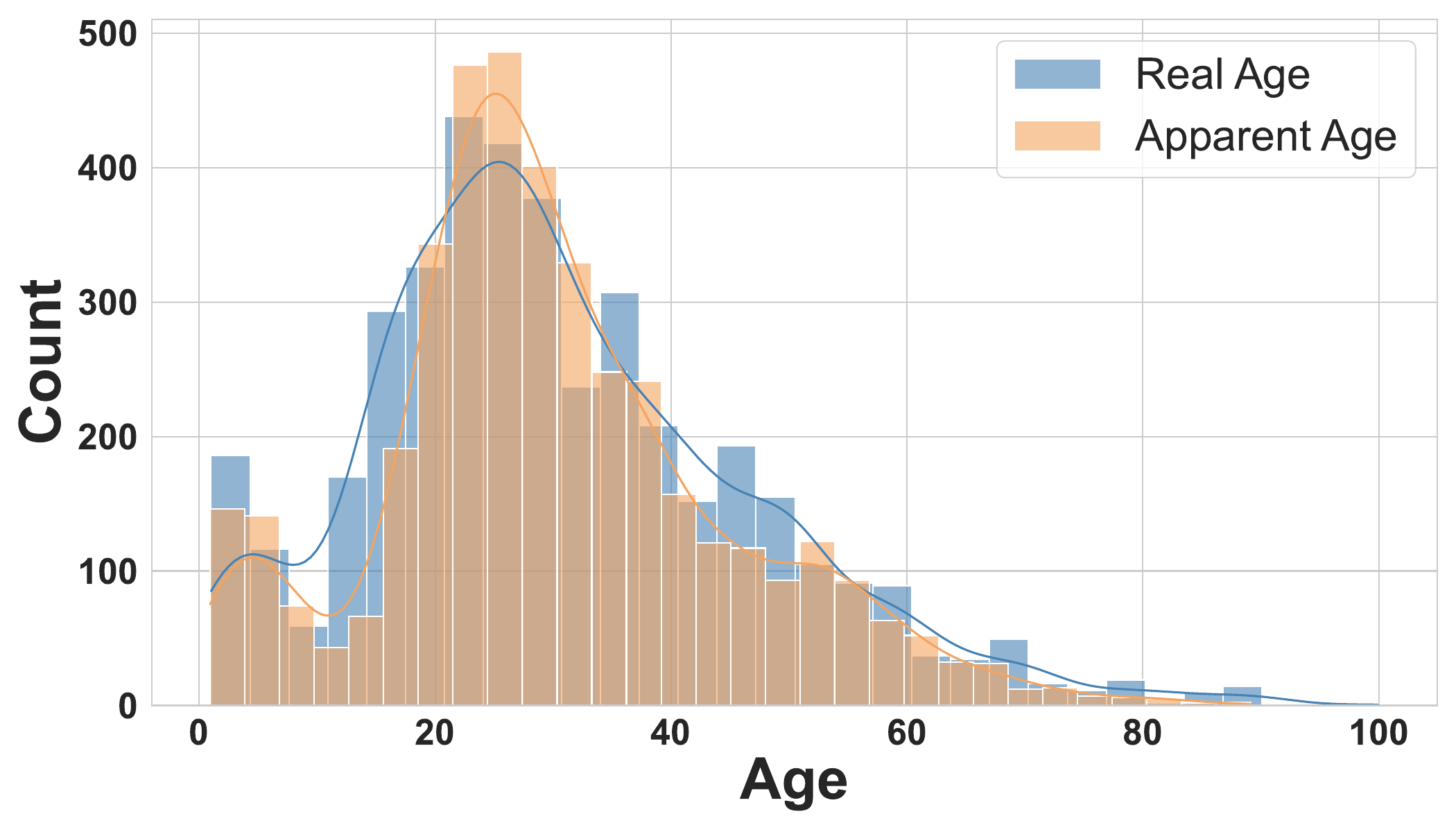}
        \label{fig:agedis}
    }
    \subfloat[Age error (Apparent - Real)]{
        \centering
        \includegraphics[width=0.32\linewidth]{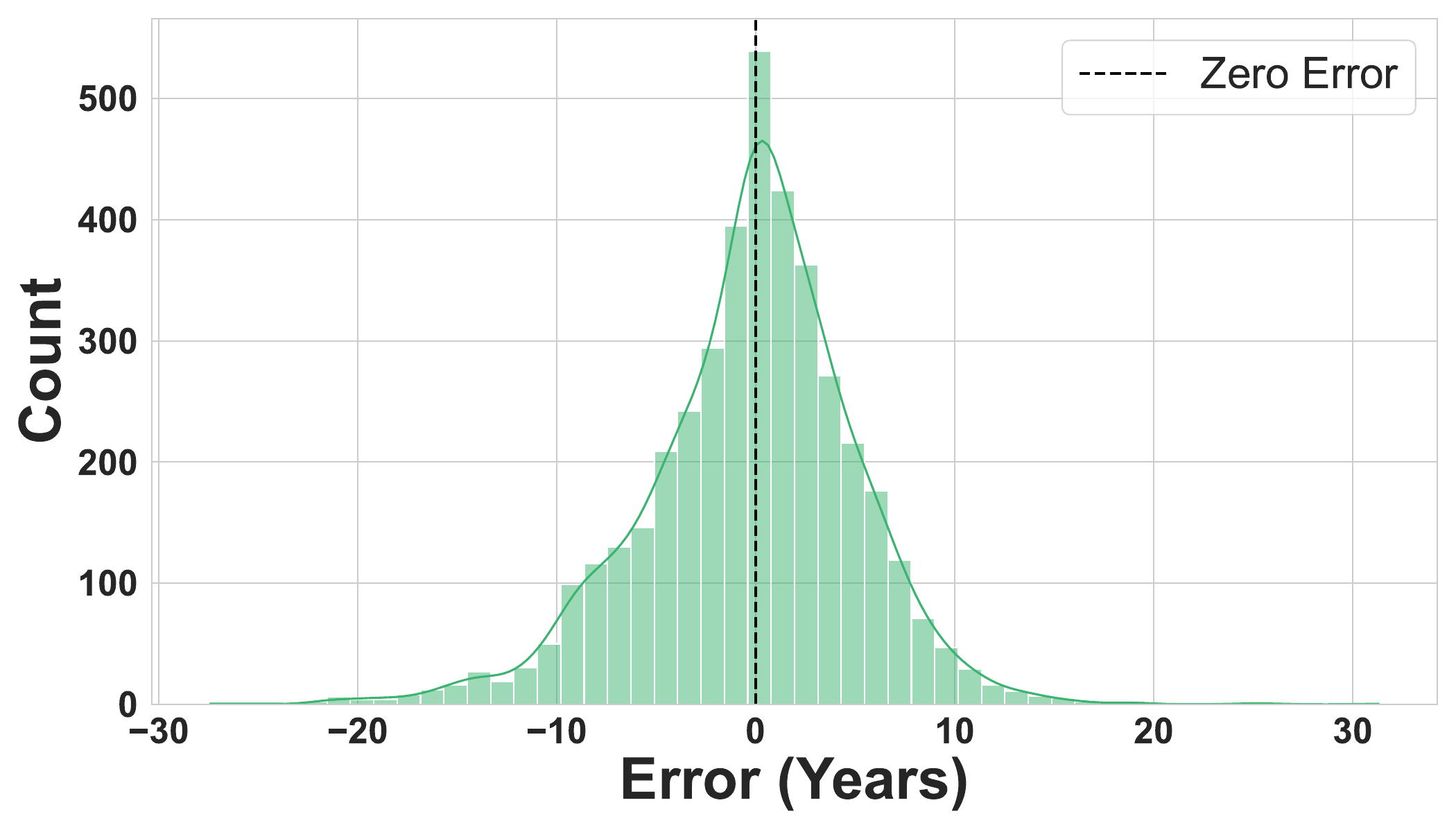}
        \label{fig:errordis}
    }
    \caption{Distribution of real and apparent age in the APPA-REAL dataset. On average, the apparent age matches the real age quite well, except for an increased uncertainty.}
\end{figure*}

\section{Background}
Previous research on Uncertainty Quantification in face age estimation has been sparse, but showing some promising results. One approach has been to group ages into age-classes, so that age estimation can be treated as a classification problem \cite{terhorst2019reliable}. This results models that produce age-class-probabilities, which can be used as a measure of uncertainty. Being selective in which classification gets used, based on a threshold on the class probability has been shown to improve age classification accuracy \cite{terhorst2019reliable}. Mixture Density Networks have been applied on age estimation based on the IMDb-Wiki dataset \cite{rothe2015}, but this dataset contains predominately Caucasian celebrities between 20 and 40 years old \cite{terhorst2019reliable}.

To evaluate whether these uncertainties are disentangled in the regression setting, we replicate the experimental approach used in classification studies: we train models on different subset sizes of the dataset \cite{dejong2025uncertainty}.

\textbf{Age Estimation From Faces}. Estimating a person’s age from their facial image is a longstanding problem in computer vision. Age estimation can be used in security or access control; AI systems estimate age to restrict access to age-sensitive content, such as alcohol purchases or gambling \cite{Burgess2021}. Brands use age estimation to tailor content and ads based on the viewer's demographic profile \cite{Burt2023} \cite{Gollapalli2023}.  

Predicting a person's apparent age is especially difficult. Unlike biological age, apparent age is subjective and often biased \cite{Pilz2022}, requiring models to approximate human perception. This makes it a more complex task than conventional regression \cite{rothe2015}. Human perception of age is affected by systematic biases, such as the regression-to-the-mean effect, which causes younger individuals to be overestimated and older individuals to be underestimated in age estimation tasks \cite{burt2007} \cite{Clifford2018}.

Additionally, external factors such as alcohol and tobacco consumption can significantly affect perceived age. Studies have shown that heavy drinking and smoking accelerate visible facial aging, making individuals appear older than their biological age \cite{Goodman2019FacialAging}.

Another challenge arises from the visual similarity within specific age ranges. Older adults, between 60 and 80 years old, tend to have less distinctive facial features that differentiate their precise age, leading to higher estimation errors \cite{Ganel2023}. Similarly, prepubescent children, under 13, display subtle differences in facial features that make age estimation in this group particularly difficult \cite{Ferguson2017}.

Despite these systematic tendencies to underestimate older adults and overestimate children, quantifying exactly how large those deviations can become is crucial. Figures \ref{fig:old} and \ref{fig:young} respectively show an 87‑year‑old and a 16‑year‑old, each annotated by the same group of raters. These examples highlight the scale and direction of human perceptual bias.

\subsection{Uncertainty Estimation}
In regression problems, we can treat model and data uncertainty as variances, and have the total predictive variance $\sigma^2_{\text{pred}}(\cdot)$ for some input $x^*$ as
\begin{equation}
\sigma^2_{\text{pred}}(x^*) = 
\underbrace{\mathbb{E}\left[\sigma^2_{\text{aleatoric}}(x^*)\right]}_{\text{Aleatoric uncertainty}} + 
\underbrace{\mathrm{Var}\left[\mu(x^*)\right]}_{\text{Epistemic uncertainty}}.
\label{eq:predictive_uncertainty}
\end{equation}
The disentanglement of the uncertainties is the separation of these two aforementioned uncertainties. %
Aleatoric uncertainty is inherent uncertainty in the data. In apparent age estimation, it arises from annotator disagreement. Mathematically, aleatoric uncertainty is expressed as the variance of the output 
$\sigma^2_{\text{aleatoric}}(x) = \mathrm{Var}[y \mid x]$.

In this study, the variance is learned alongside the prediction of the mean, enabling the model to express confidence in its outputs by explicitly estimating the aleatoric uncertainty.

On the other hand, epistemic uncertainty arises not from the data itself but from the limitations of the model. It can be caused by various problems, including small datasets, class imbalance, overfitting, underfitting, or, more broadly, the model failing to generalize. In Bayesian models, epistemic uncertainty is estimated by evaluating the extent to which the model’s predictions vary across different parameter configurations. This idea is captured in the following formula: $\mathrm{Var}[\mu(x)] = \frac{1}{T} \sum_{t=1}^{T} \mu_t^2(x) - \left( \frac{1}{T} \sum_{t=1}^{T} \mu_t(x) \right)^2$. Here, the model is run \(T\) times for the same input \(x\), each time with different dropout configurations, resulting in different predicted means \(\mu_t(x)\). The first term, \(\frac{1}{T} \sum_{t=1}^{T} \mu_t^2(x)\), is the average of the squared predictions, and the second term, \(\left( \frac{1}{T} \sum_{t=1}^{T} \mu_t(x) \right)^2\), is the square of the average prediction. Their difference yields the variance of the predictions, which quantifies the model’s uncertainty about its output.

\section{Facial Age Estimation with Uncertainty}
\textbf{Dataset}. This study uses the APPA-REAL dataset \cite{agustsson2017appareal}, which contains 7,591 images annotated with both real age and apparent age. Apparent age labels are obtained from human perception studies: each image was rated by multiple annotators, and the final apparent age corresponds to the mean of all votes. In total, the dataset comprises approximately 250,000 individual annotations, with an average of 38 votes per image, resulting in a low standard error of the mean of 0.3.

The dataset is split into 4,113 training images, 1,500 validation images, and 1,978 test images. The training set exhibits a strong class imbalance, with a high concentration of samples in the 20–40 age range (Figure~\ref{fig:agedis}). 
\begin{figure*}[t]
    \subfloat[DropConnect]{
        \includegraphics[width=0.32\linewidth]{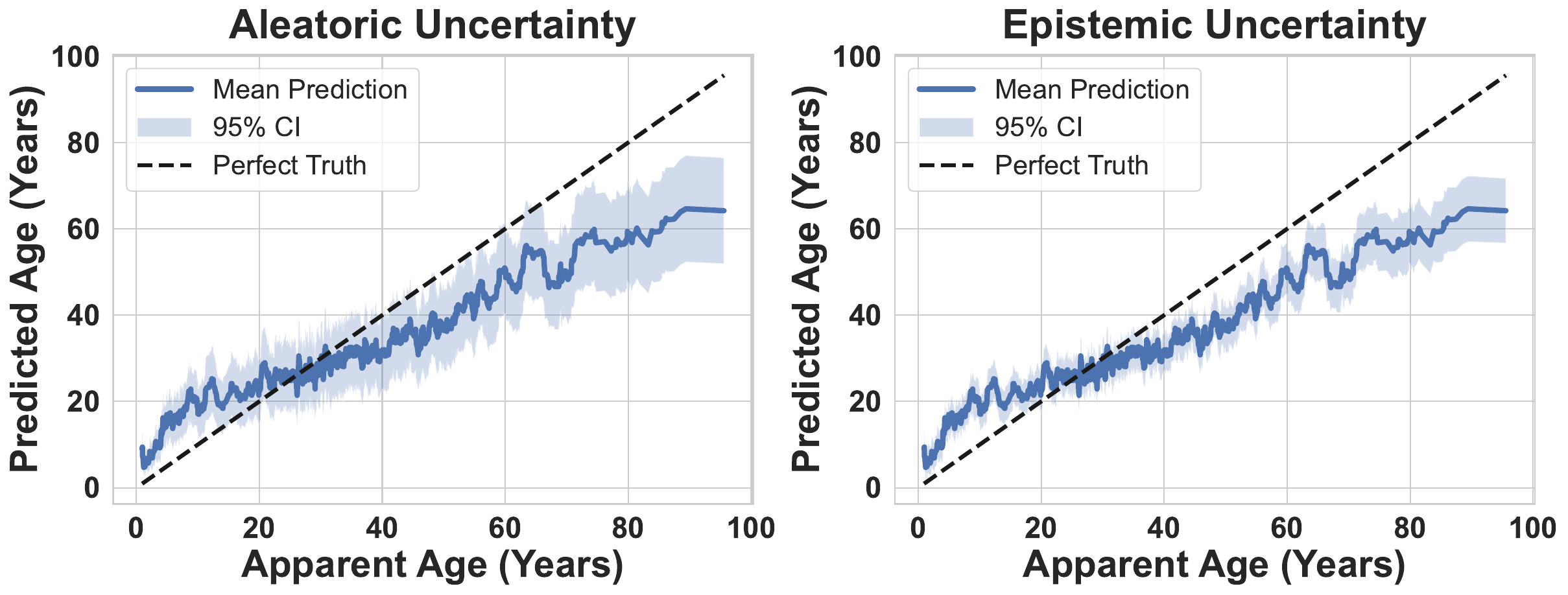}
    }
    \subfloat[Ensembles]{
        \includegraphics[width=0.32\linewidth]{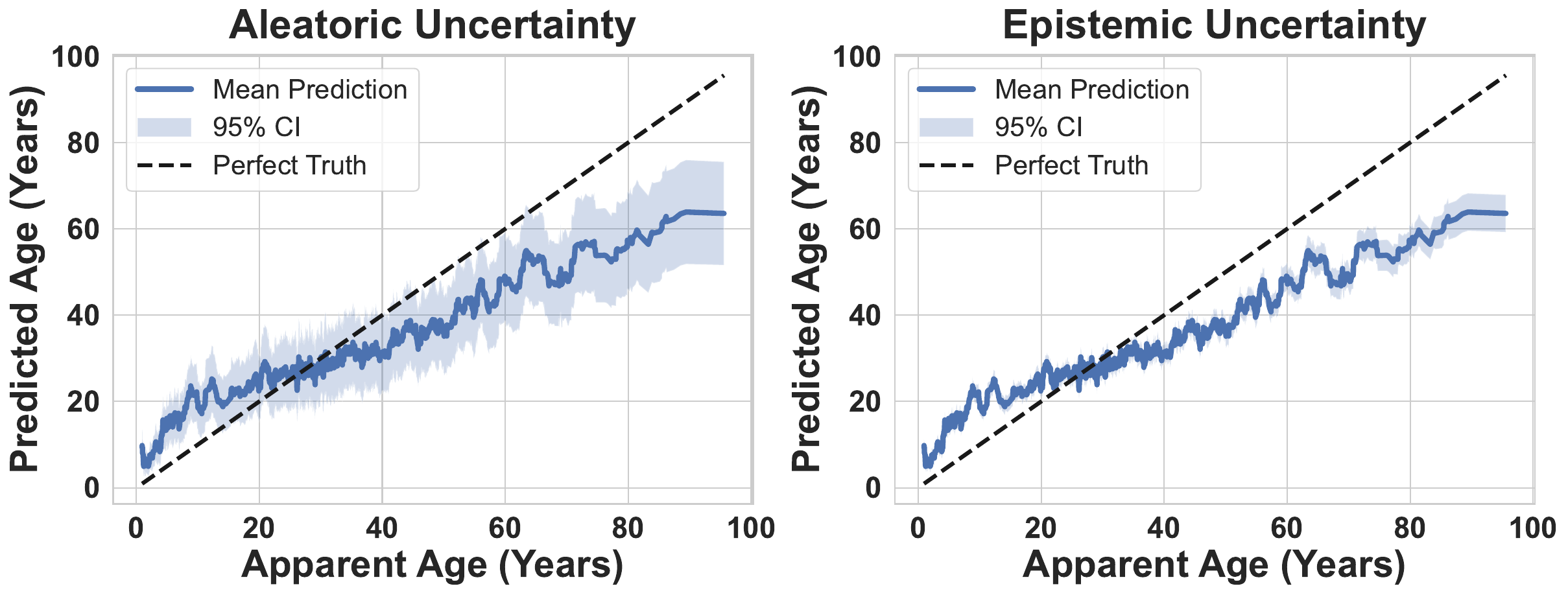}
    }
    \subfloat[Flipout]{
        \includegraphics[width=0.32\linewidth]{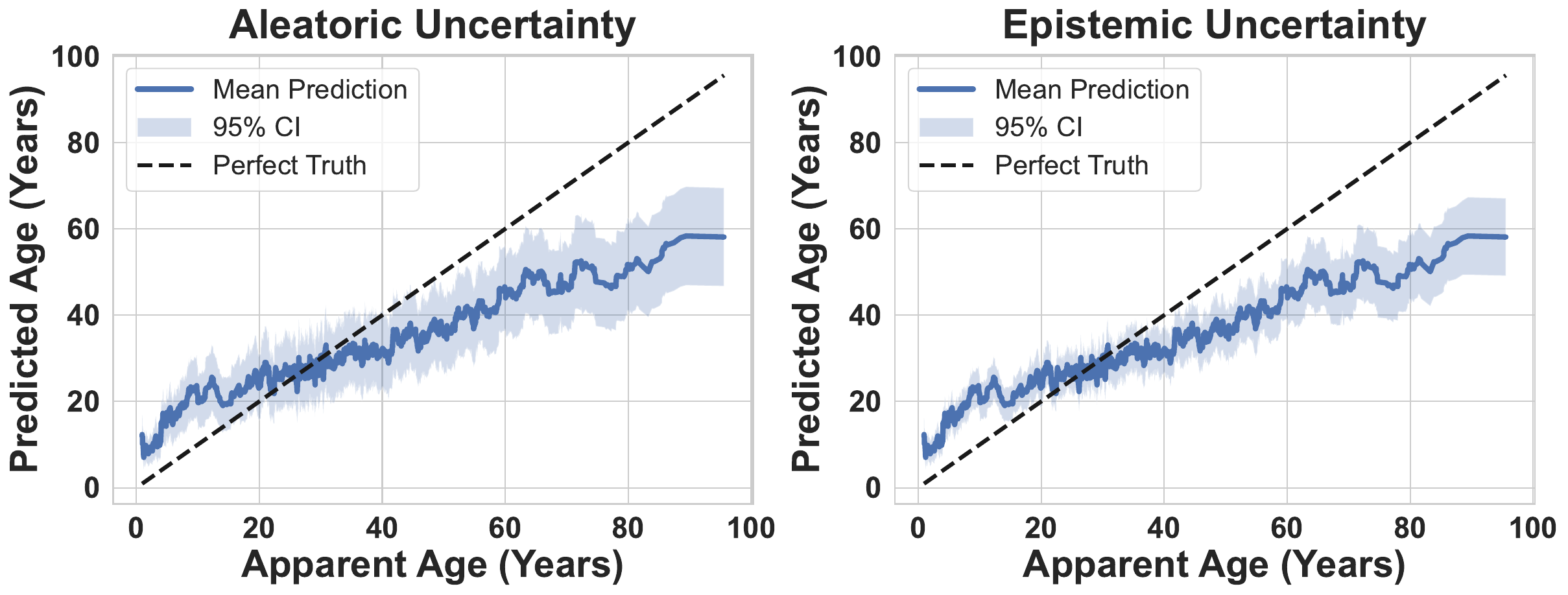}
    }
    \caption{Comparison of uncertainties for apparent age prediction, showing overestimation for younger ages ($\leq 30$) and underestimation for older ages ($>30$).}
\end{figure*}
\textbf{Preprocessing}. A custom data loading and preprocessing pipeline was implemented using TensorFlow and Keras to ensure consistency across experiments and efficient training. Since models are trained on varying proportions of the dataset, a stratified subsampling strategy is employed to preserve the original label distribution across subsets.

Images are resized to $224 \times 224$ pixels using Keras utilities. Data augmentation is applied to improve robustness and reduce overfitting, including random rotations ($\pm 10^\circ$), horizontal and vertical shifts (up to 10\%), zooming (80–120\%), horizontal flipping, and brightness variation (80–120\%). For each sample, a random augmentation strategy is applied, followed by normalization.

\textbf{Model architecture}. The proposed model is a multi-output regression architecture that jointly predicts the mean apparent age and its standard deviation. The backbone is a DenseNet-121 network \cite{huang2017densenet} pre-trained on ImageNet \cite{deng2009imagenet}, selected for its efficient feature reuse and strong gradient propagation.

Extracted features are passed to a shared regression head consisting of two fully connected layers with 512 and 256 neurons. Each layer is followed by batch normalization and Leaky ReLU activation. L2 regularization is applied to dense weights, and a dropout layer with a rate of 0.5 is used after the shared layers to improve generalization.

The network then branches into two outputs. The mean apparent age branch uses a 128-neuron dense layer with ReLU activation followed by a single-neuron ReLU output to enforce non-negative predictions. The uncertainty branch mirrors this structure but uses a Softplus activation in the final layer to ensure strictly positive standard deviation estimates.

\textbf{Training Setup}. We implemented a comprehensive training pipeline structured in three distinct phases: initial single-task training (only mean age via MSE), multitask training (both mean and std age via MSE), and fine-tuning (all layers). 

The model produces two outputs per image: mean apparent age and apparent age uncertainty, expressed as a standard deviation. Crucially, the uncertainty branch is explicitly supervised. The APPA-REAL dataset provides multiple human annotations per image; from these annotations, we compute the mean apparent age, used as the regression target for the mean branch, and the standard deviation of human votes, which reflects annotator disagreement and serves as a proxy for aleatoric uncertainty. This yields a per-image uncertainty target that allows the model to learn how uncertain humans are about a given face, rather than inferring aleatoric uncertainty indirectly.

The Adam optimizer is used throughout training with a learning rate of $2 \times 10^{-4}$, which is reduced to $1 \times 10^{-5}$ during fine-tuning.

\begin{figure*}[!t]    
    \centering
    \subfloat[DropConnect]{
        \includegraphics[width=0.32\linewidth]{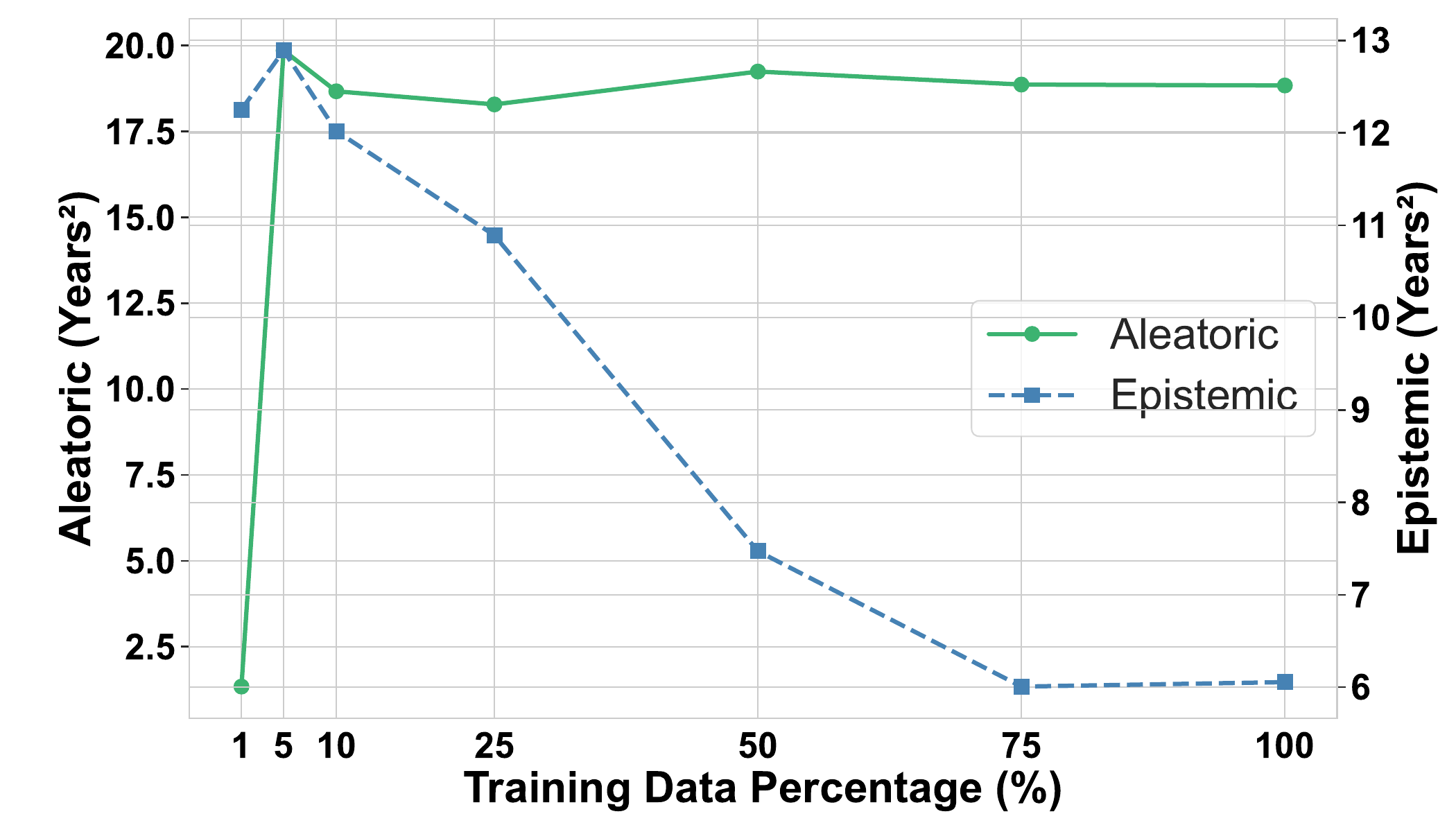}
        \label{fig:dropconplot}
    }        
    \subfloat[Ensembles]{
        \includegraphics[width=0.32\linewidth]{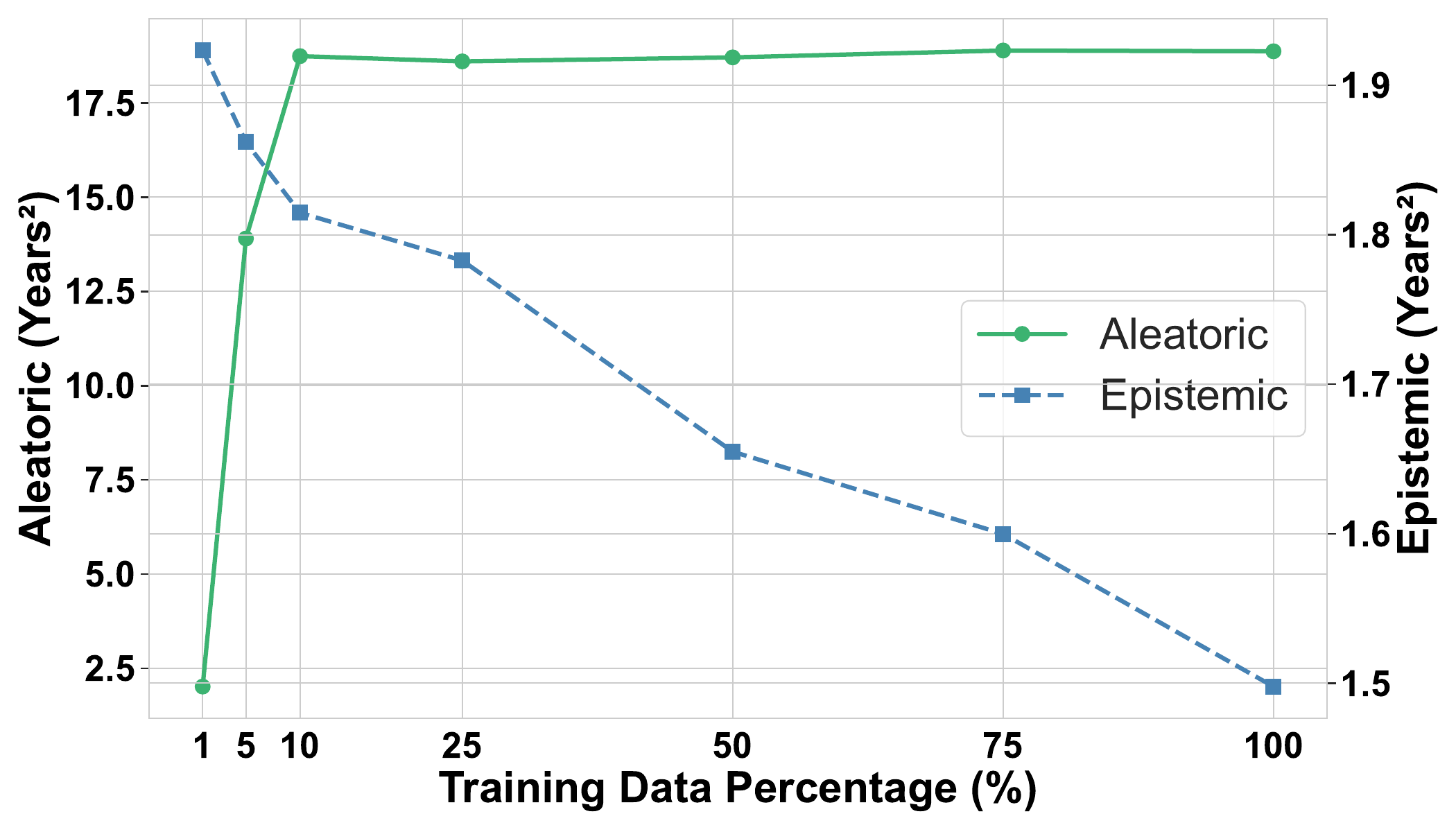}
        \label{fig:ensembleplot}
    }        
    \subfloat[Flipout]{
        \includegraphics[width=0.32\linewidth]{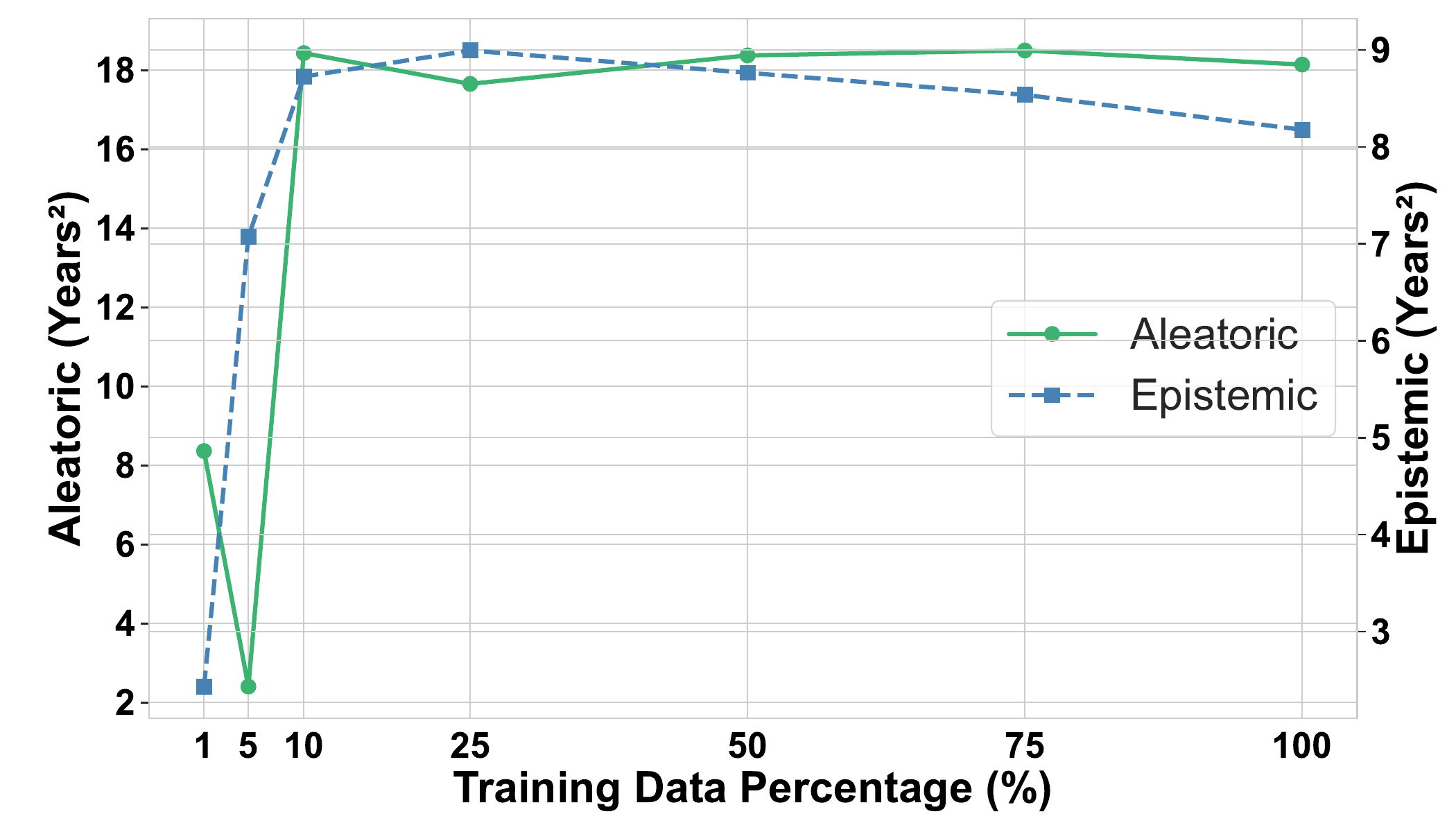}
        \label{fig:flipoutplot}
    }        
    \caption{Uncertainty trends using several uncertainty estimation methods across different training data percentages.}   
\end{figure*}

\textbf{Uncertainty Disentanglement}. To disentangle aleatoric and epistemic uncertainty, four methods are evaluated: MC-Dropout \cite{gal2016dropout}, MC-DropConnect \cite{mobiny2021dropconnect}, Flipout \cite{wen2018flipout}, and Deep Ensembles \cite{lakshminarayanan2017simple}. These approaches are inspired by Bayesian Neural Networks (BNNs), which model uncertainty by placing distributions over network weights \cite{blundell2015weight}.

\textbf{Experimental Setup}. To find how the epistemic and aleatoric uncertainty behave with respect to dataset size, we trained the models on seven different percentages of the dataset. These percentages are: 1\%, 5\%, 10\%, 25\%, 50\%, 75\%, 100\%; they were chosen based on the study by \cite{dejong2025uncertainty}.  Each subset has the same distribution of the labels, so the only difference between the subsets is the number of samples. For example, the smallest subset (1\% of the training set) has only 40 samples. 

\section{Results}

Figures~\ref{fig:dropconplot}–\ref{fig:flipoutplot} summarize the estimated aleatoric and epistemic uncertainties across different training data percentages for the three uncertainty estimation methods.

Across all methods, aleatoric uncertainty remains largely stable as dataset size increases, while epistemic uncertainty generally decreases, which is the expected behavior.

\textbf{Uncertainty Trends Across Dataset Sizes} Figures~\ref{fig:dropconplot}–\ref{fig:flipoutplot} reports the corresponding aleatoric and epistemic uncertainties for training data percentages ranging from 1\% to 100\%. Training on 1\% of the dataset corresponds to approximately 40 samples and resulted in severe underfitting. Consequently, uncertainty estimates at this level are unreliable and excluded from interpretation.

\begin{figure*}[!t]
    \centering
    \includegraphics[width=0.24\linewidth]{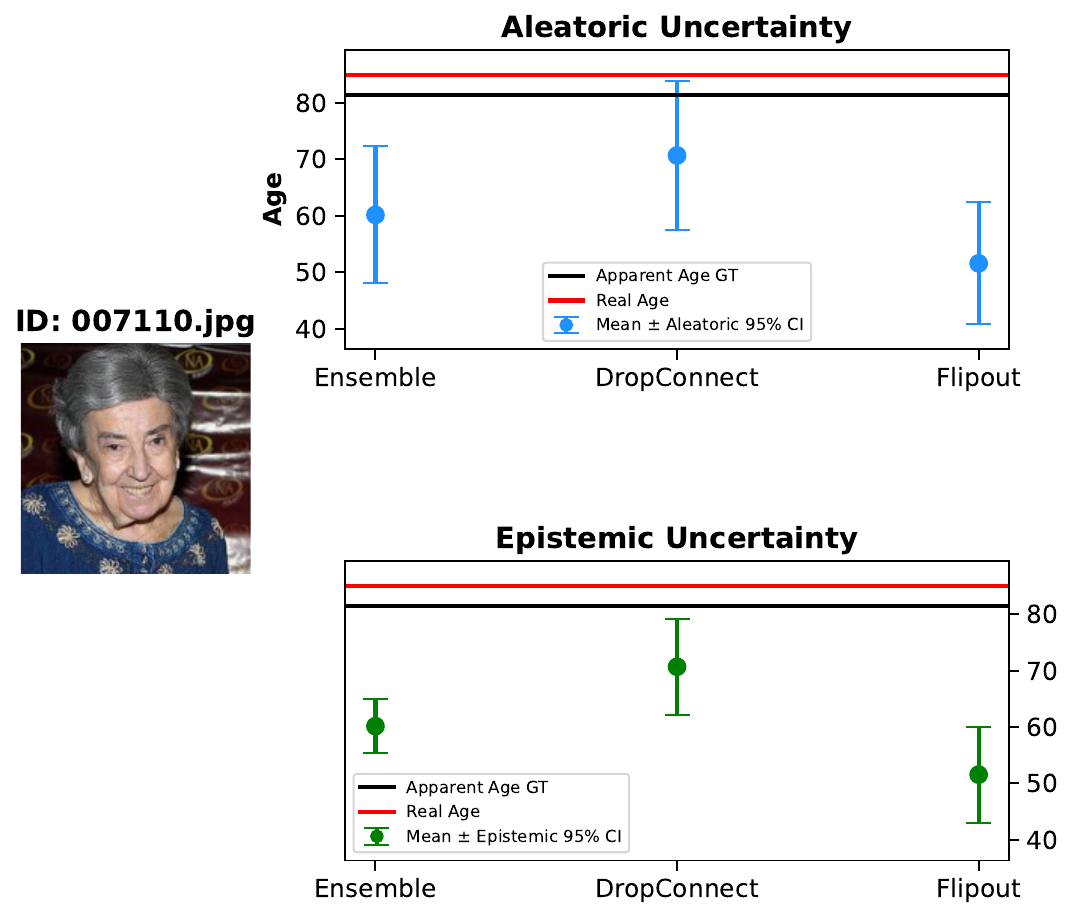}
    \includegraphics[width=0.24\linewidth]{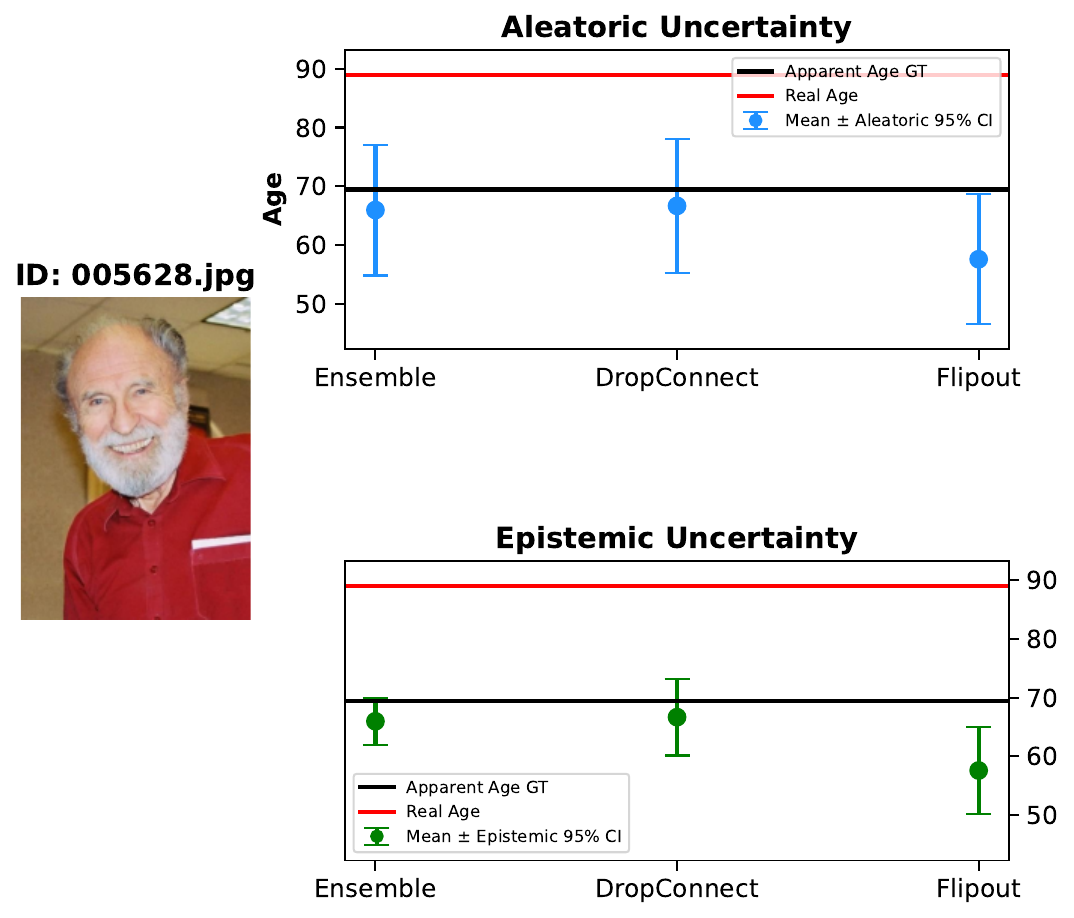}    
    \includegraphics[width=0.24\linewidth]{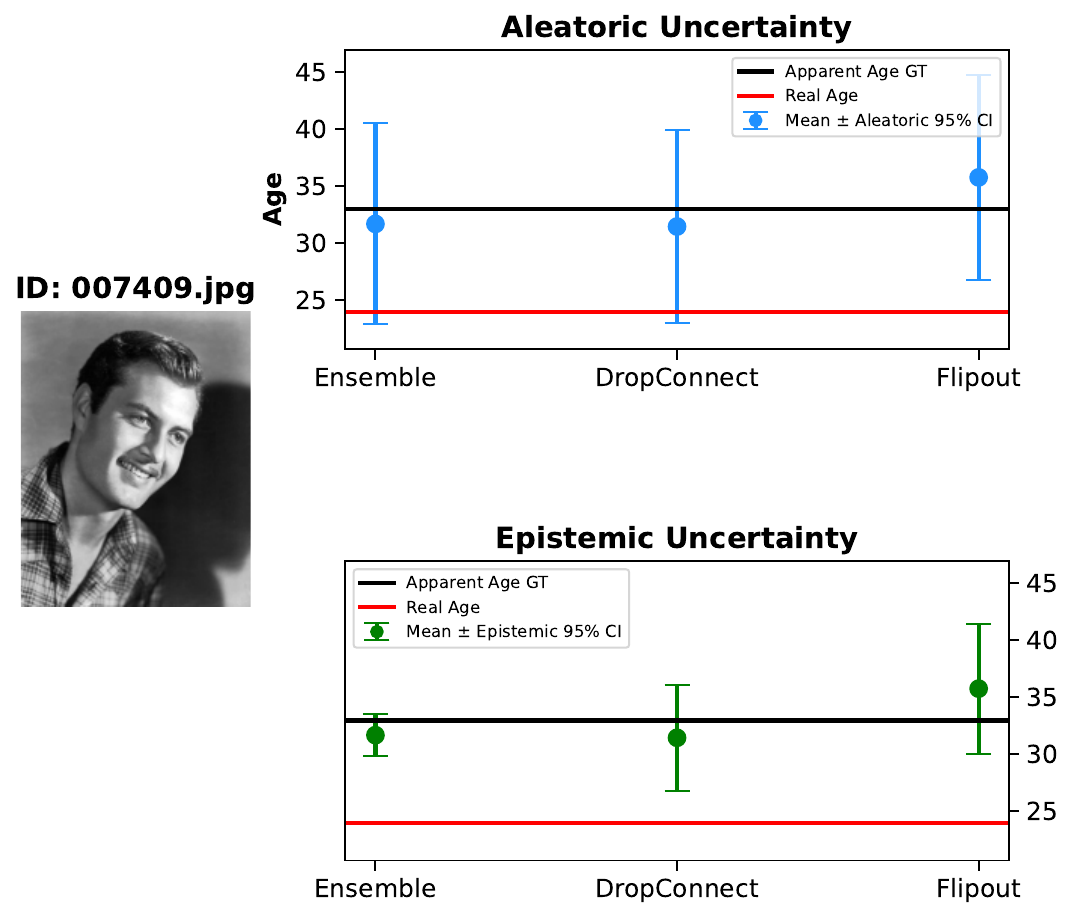}
    \includegraphics[width=0.24\linewidth]{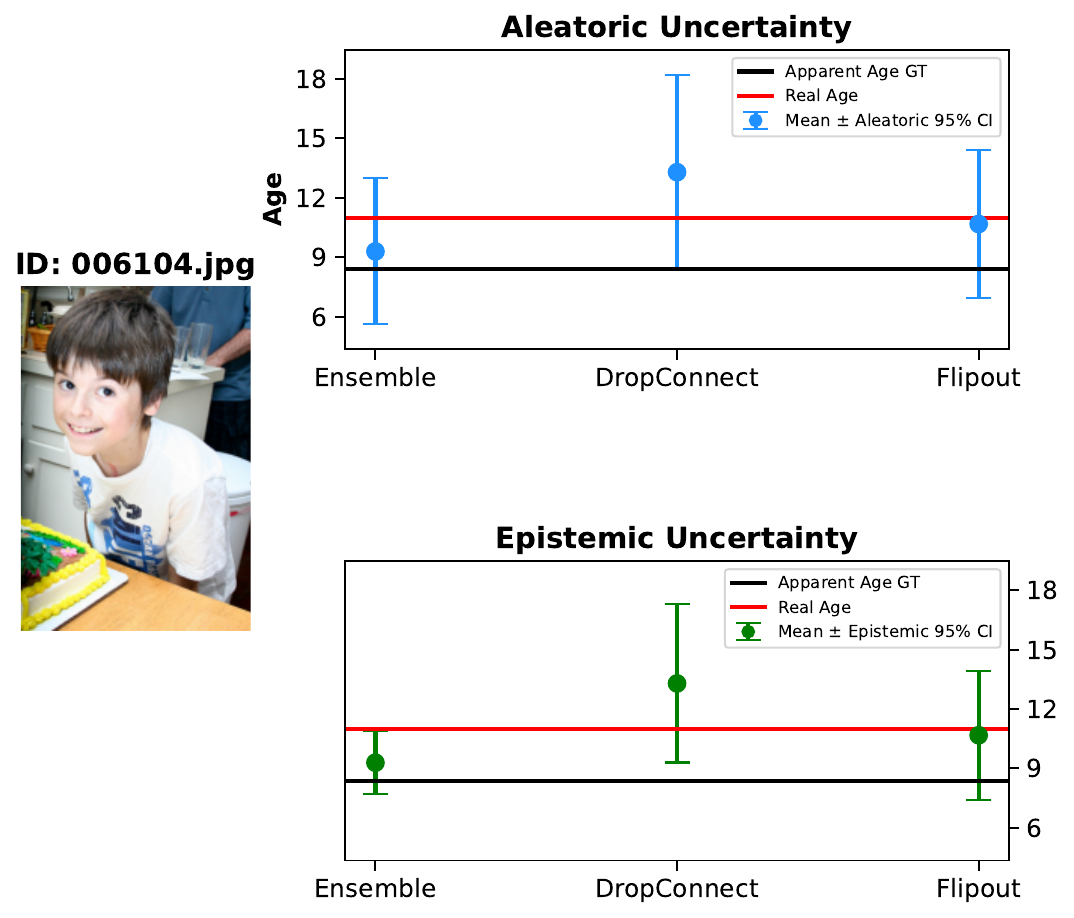}
    \caption{Uncertainty estimates for several test images. For each: left, the facial image; right, corresponding model predictions with 95\% confidence intervals for aleatoric (blue) and epistemic (green) uncertainty. The black horizontal line denotes the ground truth apparent age (GT); the red line denotes the real age. Error bars reflect $\pm$1.96 standard deviations (95\% CI).}
    \label{fig:uncertainty_multiple1}
\end{figure*}
Across all meaningful dataset sizes, aleatoric uncertainty remains stable at approximately 18 years squared, reinforcing its data-inherent nature. In contrast, epistemic uncertainty is consistently lower at 100\% training data than at smaller data fractions, with the steepest decline observed for DropConnect.

\begin{figure*}[t]
    \centering
    \subfloat[Mean Absolute Error]{
        \includegraphics[width=0.32\linewidth]{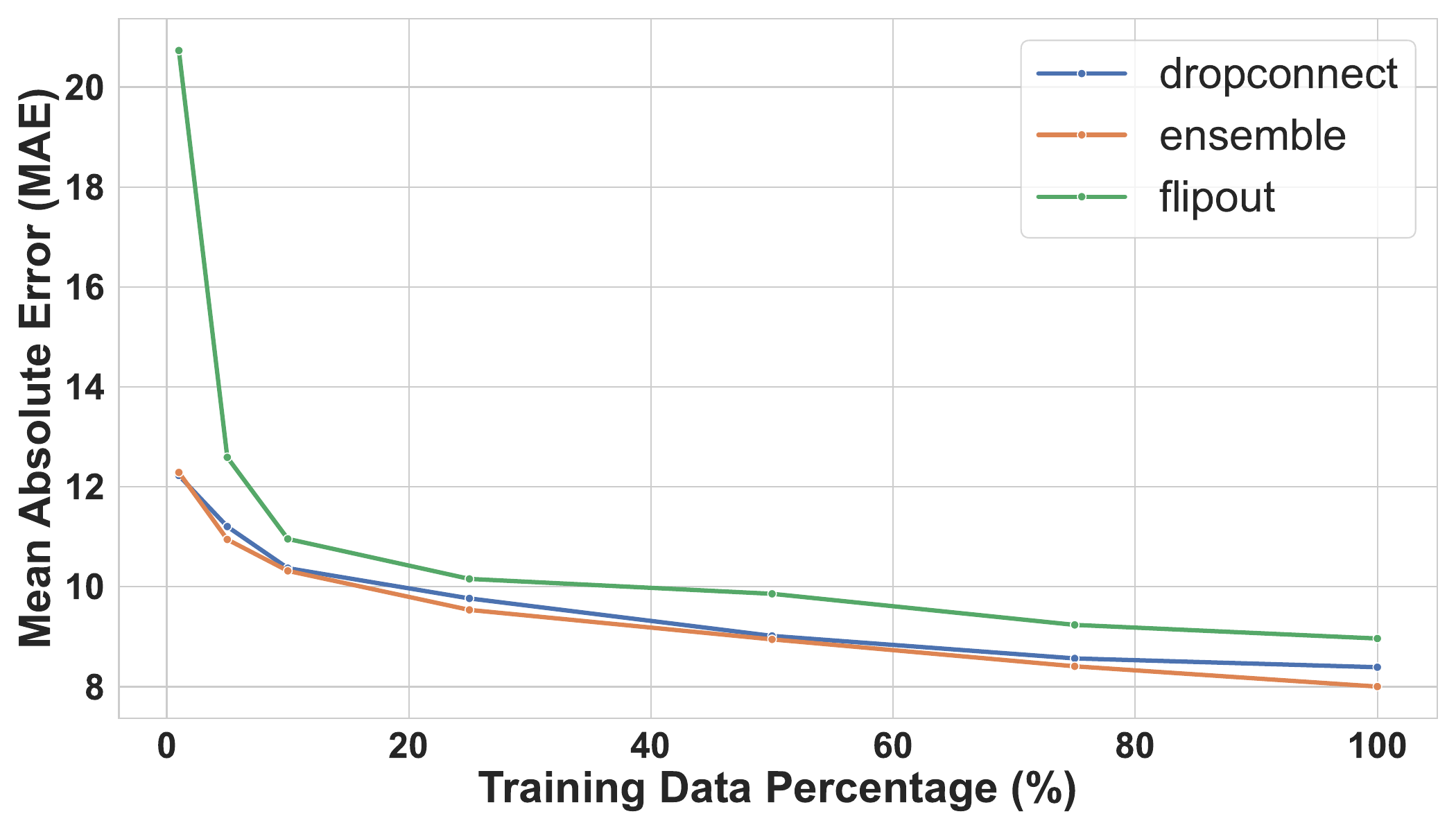}
        \label{fig:mae}
    }
    \subfloat[Root Mean Squared Error]{
        \includegraphics[width=0.32\linewidth]{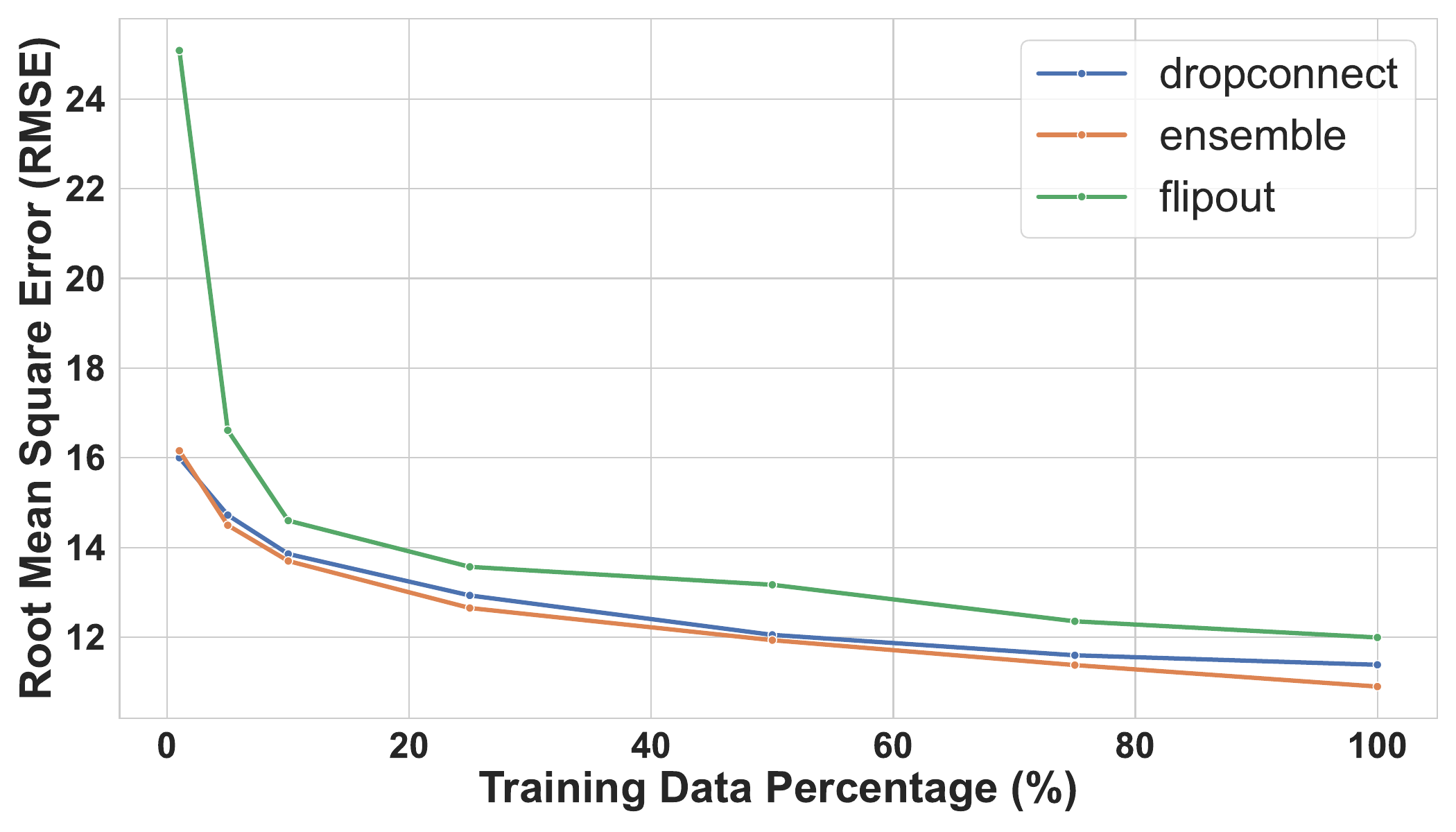}
        \label{fig:rmse}
    }
    \subfloat[$R^2$ score]{
        \includegraphics[width=0.32\linewidth]{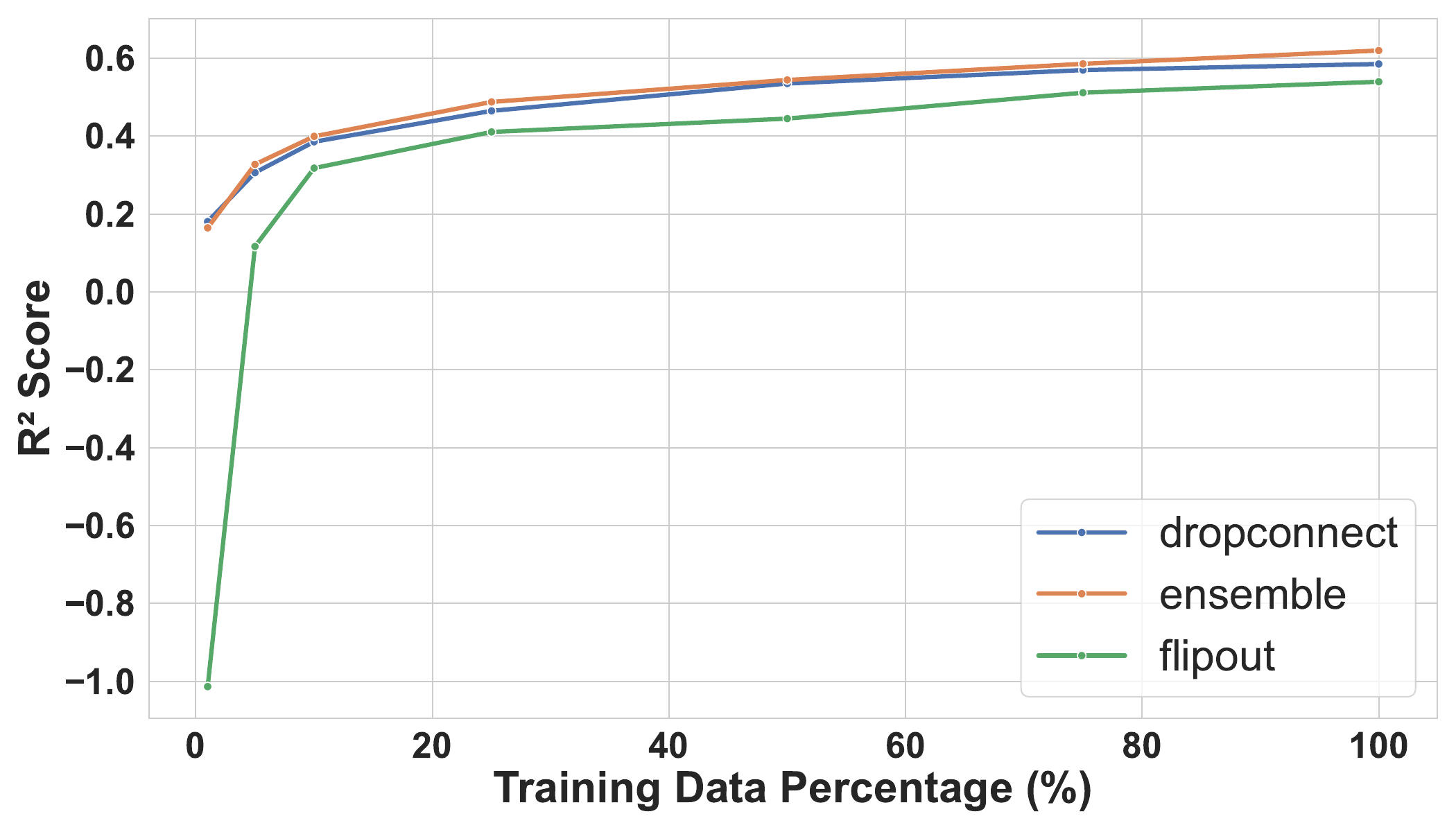}
        \label{fig:r2}
    }
    \caption{Comparison of metrics across different training set sizes.}
\end{figure*}

\textbf{Prediction Error Across Dataset Sizes}. In addition to uncertainty estimation, we evaluate predictive accuracy using MAE, RMSE, and $R^2$ across all dataset sizes and methods. Figure~\ref{fig:mae} shows that MAE decreases steadily with increasing data for all methods. DropConnect reduces its MAE from over 11.2 years at 5\% data to 8.3 years at full data. Deep Ensembles achieve the best overall performance with a final MAE of 8.0 years, while Flipout converges more slowly, reaching 8.9 years at 100\%.

The relative performance is further illustrated in Figure~\ref{fig:r2}. At full data, Deep Ensembles achieve an $R^2$ score of 0.61, followed by DropConnect at 0.58 and Flipout at 0.53. These results confirm that predictive accuracy improves alongside increasing data availability.

Overall, all models benefit from larger training datasets, but Deep Ensembles provide the strongest raw predictive performance. Importantly, improvements in accuracy are accompanied by reductions in epistemic uncertainty, indicating that increased confidence coincides with improved prediction quality as data availability grows.

\textbf{Qualitative Comparison of Uncertainty Estimates on Sample Inputs}. To complement quantitative results, we qualitatively examine uncertainty behavior using four representative samples from the APPA-REAL test set spanning different age groups in Figure~\ref{fig:uncertainty_multiple1}. %

Elderly subjects exhibit consistently high aleatoric uncertainty, indicating substantial annotator disagreement in estimating apparent age for older faces. Epistemic uncertainty is also elevated in this group, particularly for Flipout, reflecting limited model familiarity with underrepresented age ranges. DropConnect and Ensemble methods show comparatively narrower epistemic intervals, suggesting improved robustness in data-scarce regimes. In contrast, the child sample displays high aleatoric but low epistemic uncertainty, implying that while human perception of child age remains ambiguous, the model has sufficient exposure to young faces to generalize effectively. The young adult subject demonstrates low uncertainty in both components, corresponding to strong annotator consensus and high data representation for this demographic.

\begin{figure*}[t]
    \centering
    \includegraphics[width=\linewidth]{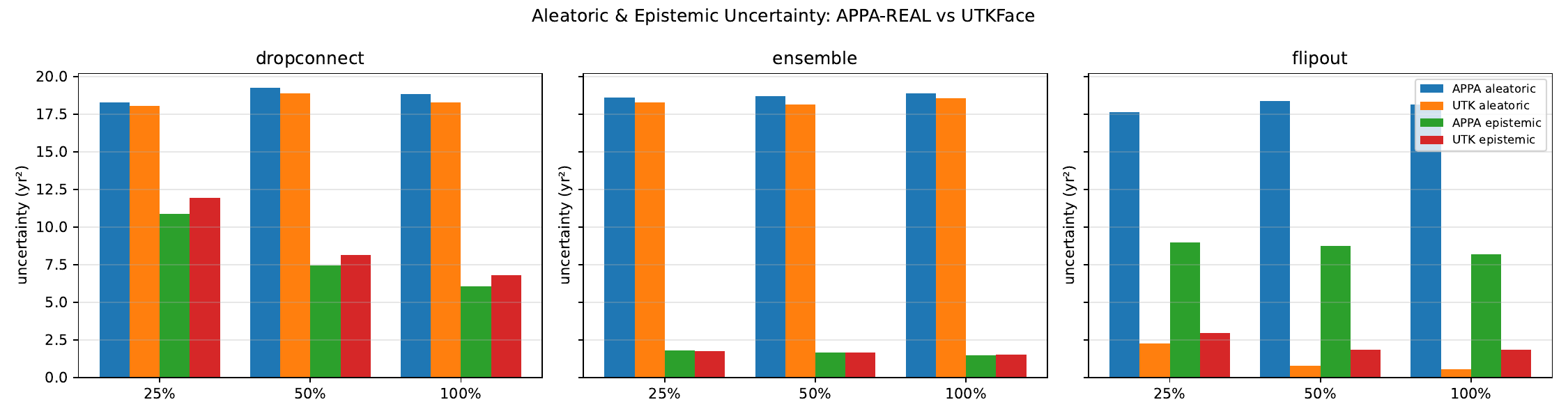}
    \caption{Comparison of uncertainties between APPA vs UTK test sets. DropConnect and Ensemble aleatoric uncertainties do not vary, but epistemic uncertainties slightly increase, and Flipout severely underpreforms.}
    \label{fig:utk_comparison}
\end{figure*}

\textbf{Generalization}.  We also evaluate our models on the UTKFace dataset for generalization \cite{zhifei2017cvpr}, while mean absolute errors increase from around 8 to 16 in DropConnect and Ensembles, and 10 to 30 for Flipout. Both types of uncertainty shown in Figure \ref{fig:utk_comparison} are relatively similar for DropConnect and ensembles with a slight increase of Epistemic uncertainty, while they both significantly drop for Flipout.  We interpret these results as our model's uncertainty being able to generalize to other datasets like UTKFace with different data distribution, and epistemic uncertainty does reflect a degradation in performance.

\section{Discussion}

\textbf{Interpretation of Results}. The empirical trends observed are consistent with theoretical definitions of uncertainty. Aleatoric uncertainty remained largely invariant across different training set sizes, confirming its data-inherent nature, as previously reported by \cite{kendall2017uncertainties}. Across models and dataset sizes, aleatoric variance stabilized at approximately 18 years squared, corresponding to a standard deviation of 4.2 years, which is plausible given the subjective difficulty of estimating perceived age.

Epistemic uncertainty, in contrast, exhibited the expected inverse relationship with training data size. DropConnect demonstrated a pronounced reduction, decreasing from approximately 12.9 years at 5\% of the data to 6.0 years when trained on the full dataset, reflecting increased model confidence with broader data exposure. Deep Ensembles showed consistently low epistemic uncertainty and reduced sensitivity to data size, likely due to robustness induced by model diversity. Flipout displayed less stable behavior, with non-monotonic fluctuations, particularly at intermediate dataset sizes, suggesting potential instability in its variational inference approximations.

\textbf{Relation to Existing Literature}. These findings contrast with prior classification studies \cite{dejong2025uncertainty}, where strong correlations between aleatoric and epistemic uncertainty hindered successful disentanglement. In the regression setting examined here, the two uncertainty types were substantially more separable, both conceptually and empirically, suggesting that regression tasks may be better suited for uncertainty disentanglement using current approximate Bayesian methods.

The consistently low epistemic uncertainty observed in Deep Ensembles, even under limited data conditions, aligns with previous work by \cite{lakshminarayanan2017simple}, reinforcing the view that ensembles act as strong non-Bayesian uncertainty approximators. However, this same robustness may mask sensitivity to data scarcity, potentially leading to underestimation of epistemic uncertainty in low-data regimes.

\textbf{Scalability with More Data}. A common issue with any machine learning model, is the training data and the question if more data will improve the model's predictions. Results in Figures \ref{fig:mae} and \ref{fig:rmse} evaluate standard regression metrics across dataset size, and from these results we can interpolate that the model can continue improving with more data, as the MAE/RMSE slope is negative and not constant to indicate possible diminishing gains.

\textbf{Limitations}. A key limitation of this study is the reliance on very small training subsets. Models trained on 1\% of the data (approximately 40 samples) exhibited clear underfitting, resulting in unreliable predictions and uncertainty estimates; consequently, these results were excluded from the final analysis. Even at larger subset sizes (5\% and 10\%), Flipout showed irregular and non-monotonic epistemic uncertainty trends, including temporary increases despite additional training data.

The exclusive use of the APPA-REAL dataset also limits generalizability. Although the dataset provides high-quality annotations with low standard error and is well-suited for apparent age estimation, its age distribution is skewed toward individuals aged 20–40, with underrepresentation of children and elderly adults. This imbalance may impair model generalization and affect uncertainty estimates in sparsely represented age ranges.

Another limitation concerns architectural homogeneity. All experiments employed DenseNet121 as the backbone architecture due to its efficiency and strong performance. However, uncertainty behavior may depend on architectural properties such as depth, feature reuse, or inductive biases. It therefore remains unclear whether similar disentanglement patterns would emerge with alternative architectures, such as ResNet \cite{he2016deep} or Vision Transformers \cite{dosovitskiy2021an}.

\textbf{Future Work}. Future research should address these limitations while extending the scope of uncertainty-aware regression modeling. One promising direction is the development of hybrid uncertainty estimation approaches that combine Bayesian and non-Bayesian methods, potentially reducing variance in epistemic estimates while preserving robustness. Ensemble-based Bayesian approximations may offer a balanced trade-off between calibration and stability.

\section{Conclusions}\label{sec:conclusions}

Our results demonstrate that uncertainty disentanglement is not only feasible in regression but also more stable and interpretable compared to classification. Aleatoric uncertainty remained consistent across all model types and dataset sizes, confirming its origin in data-inherent variability. In contrast, epistemic uncertainty exhibited a clear inverse relationship with training dataset size, decreasing as more data became available. Thus, it validates its dependence on model knowledge and data exposure.

Among the methods evaluated, Deep Ensembles provided the best overall predictive performance and maintained low uncertainty levels, although they appeared less sensitive to changes in training size. DropConnect, although slightly less accurate, showed the most responsive and interpretable epistemic trends, making it a strong candidate for applications that prioritize model confidence estimation. Flipout, on the other hand, showed less stable behavior, particularly at low-data regimes, highlighting limitations in its variational approximation stability.

These findings reinforce the viability of using approximate Bayesian methods to quantify and separate predictive uncertainties in regression tasks. %
As facial age estimation becomes increasingly integrated into high-stakes applications, our study highlights the value of models that can not only predict but also explain their uncertainty.

\bibliographystyle{splncs04}
\bibliography{literature_bachelorthesis}

\begin{thebibliography}{10}
\providecommand{\url}[1]{\texttt{#1}}
\providecommand{\urlprefix}{URL }
\providecommand{\doi}[1]{https://doi.org/#1}

\bibitem{agustsson2017appareal}
Agustsson, E., Timofte, R., Escalera, S., Baro, X., Guyon, I., Rothe, R.:
  Apparent and real age estimation in still images with deep residual
  regressors on {APPA-REAL} database. In: 12th IEEE International Conference
  and Workshops on Automatic Face and Gesture Recognition (FG). IEEE (2017)

\bibitem{agustsson2017appa}
Agustsson, E., Uijlings, J., Popovic, Z., Escalera, S., van Gool, L.: Apparent
  and real age estimation in the wild. In: Proceedings of the IEEE Conference
  on Computer Vision and Pattern Recognition Workshops (CVPRW) (2017)

\bibitem{blundell2015weight}
Blundell, C., Cornebise, J., Kavukcuoglu, K., Wierstra, D.: Weight uncertainty
  in neural networks. In: Proceedings of the 32nd International Conference on
  Machine Learning (ICML). pp. 1613--1622 (2015)

\bibitem{Burgess2021}
Burgess, M.: This {AI} predicts how old children are. can it keep them safe?
  WIRED  (October 2021), accessed: 2025-06-19

\bibitem{Burt2023}
Burt, C.: Anonymized demographic data, targeted advertising added to {ITL}’s
  age estimation. Biometric Update  (February 2023)

\bibitem{burt2007}
Burt, D.M., Perrett, D.I.: Perception of age in adult caucasian male faces: The
  effects of facial hair and facial adiposity. Proceedings of the Royal Society
  B: Biological Sciences  \textbf{274}(1611),  2335--2340 (1995)

\bibitem{Clifford2018}
Clifford, C.W.G., Watson, T.L., White, D.: Two sources of bias explain errors
  in facial age estimation. Trends in Cognitive Sciences  \textbf{22}(11),
  820--829 (2018). \doi{10.1016/j.tics.2018.06.004}

\bibitem{deng2009imagenet}
Deng, J., Dong, W., Socher, R., Li, L.J., Li, K., Fei-Fei, L.: Imagenet: A
  large-scale hierarchical image database. In: 2009 IEEE Conference on Computer
  Vision and Pattern Recognition. pp. 248--255. IEEE (2009)

\bibitem{dosovitskiy2021an}
Dosovitskiy, A., Beyer, L., Kolesnikov, A., Weissenborn, D., Zhai, X.,
  Unterthiner, T., Dehghani, M., Minderer, M., Heigold, G., Gelly, S.,
  Uszkoreit, J., Houlsby, N.: An image is worth 16x16 words: Transformers for
  image recognition at scale. International Conference on Learning
  Representations (ICLR)  (2021)

\bibitem{escalera2015chalearn}
Escalera, S., Fabian, J., Bar{\'o}, X., et~al.: {ChaLearn} looking at people
  2015: Apparent age and cultural event recognition. In: Proceedings of the
  IEEE International Conference on Computer Vision Workshops (ICCVW). pp.~1--9
  (2015)

\bibitem{escalera2017chalearn}
Escalera, S., Fabian, J., Pardo, P., et~al.: {ChaLearn} looking at people 2016:
  A round-up. Pattern Recognition Letters  \textbf{72},  77--85 (2017)

\bibitem{Ferguson2017}
Ferguson, E., Wilkinson, C.: Juvenile age estimation from facial images.
  Forensic Science International  \textbf{272},  175--180 (2017).
  \doi{10.1016/j.forsciint.2016.12.013}

\bibitem{fu2010survey}
Fu, Y., Guo, G., Huang, T.S.: Age synthesis and estimation via faces: A survey.
  IEEE Transactions on Pattern Analysis and Machine Intelligence
  \textbf{32}(11),  1955--1976 (2010)

\bibitem{gal2016dropout}
Gal, Y., Ghahramani, Z.: Dropout as a bayesian approximation: Representing
  model uncertainty in deep learning. In: Proceedings of The 33rd International
  Conference on Machine Learning. vol.~48, pp. 1050--1059. PMLR (2016)

\bibitem{Ganel2023}
Ganel, T., Sofer, C., Goodale, M.A.: Biases in human perception of facial age
  are present and more exaggerated in current {AI} technology. Scientific
  Reports  \textbf{13}(1), ~69 (2023). \doi{10.1038/s41598-022-27093-0}

\bibitem{gawlikowski2021survey}
Gawlikowski, J., Tassi, C.R.N., Ali, M., Lee, J., Humt, M., Feng, J., Kruspe,
  A., Triebel, R., Jung, P., Roscher, R., et~al.: A survey of uncertainty in
  deep neural networks. Artificial Intelligence Review  \textbf{56}(Suppl 1),
  1513--1589 (2021)

\bibitem{Gollapalli2023}
Gollapalli, M., Rahman, A.u., Youldash, M., Alomari, D., Alismail, S.,
  Khawaher, F., Alkhadair, A., Aljubran, F., Alzannan, R., Alkhulaifi, D.,
  Mahmud, M.: Machine learning approach to users’ age prediction: A telecom
  company case study in {Saudi Arabia}. Mathematical Modelling of Engineering
  Problems  \textbf{10}(5),  1619--1629 (2023). \doi{10.18280/mmep.100512}

\bibitem{Goodman2019FacialAging}
Goodman, G.D., Kaufman, J., Day, D., Weiss, R., Kawata, A.K., Garcia, J.K.,
  Santangelo, S., Gallagher, C.J.: Impact of smoking and alcohol use on facial
  aging in women: Results of a large multinational, multiracial,
  cross-sectional survey. Journal of Clinical and Aesthetic Dermatology
  \textbf{12}(8),  28--39 (aug 2019)

\bibitem{he2016deep}
He, K., Zhang, X., Ren, S., Sun, J.: Deep residual learning for image
  recognition. In: Proceedings of the IEEE Conference on Computer Vision and
  Pattern Recognition (CVPR). pp. 770--778 (2016). \doi{10.1109/CVPR.2016.90}

\bibitem{he2023survey}
He, W., Jiang, Z., Xiao, T., Xu, Z., Li, Y.: A survey on uncertainty
  quantification methods for deep learning. ACM Computing Surveys  (2023)

\bibitem{huang2017densenet}
Huang, G., Liu, Z., Van Der~Maaten, L., Weinberger, K.Q.: Densely connected
  convolutional networks. In: Proceedings of the IEEE Conference on Computer
  Vision and Pattern Recognition (CVPR). pp. 4700--4708 (2017)

\bibitem{H_llermeier_2021}
Hüllermeier, E., Waegeman, W.: Aleatoric and epistemic uncertainty in machine
  learning: An introduction to concepts and methods. Machine Learning
  \textbf{110}(3),  457--506 (2021). \doi{10.1007/s10994-021-05946-3}

\bibitem{dejong2025uncertainty}
de~Jong, I.P., Sburlea, A.I., Sabatelli, M., Valdenegro-Toro, M.: Measuring
  uncertainty disentanglement error in classification. arXiv preprint
  arXiv:2408.12175  (jun 2025)

\bibitem{kendall2017uncertainties}
Kendall, A., Gal, Y.: What uncertainties do we need in bayesian deep learning
  for computer vision? In: Advances in Neural Information Processing Systems
  (NeurIPS). vol.~30, pp. 5574--5584 (2017)

\bibitem{lakshminarayanan2017simple}
Lakshminarayanan, B., Pritzel, A., Blundell, C.: Simple and scalable predictive
  uncertainty estimation using deep ensembles. In: Advances in Neural
  Information Processing Systems (NeurIPS) (2017)

\bibitem{mckinsey2024aisurvey}
McKinsey: Global {AI} survey: {AI} proves its worth, but few scale impact
  (2024), accessed: 2025-07-24

\bibitem{mobiny2021dropconnect}
Mobiny, A., Yuan, P., Moulik, S.K., Garg, N., Wu, C.C., Van~Nguyen, H.:
  Dropconnect is effective in modeling uncertainty of bayesian deep networks.
  Scientific reports  \textbf{11}(1), ~5458 (2021)

\bibitem{mucsanyi2024benchmarking}
Mucs{\'a}nyi, B., Kirchhof, M., Oh, S.J.: Benchmarking uncertainty
  disentanglement: Specialized uncertainties for specialized tasks. Advances in
  neural information processing systems  \textbf{37},  50972--51038 (2024)

\bibitem{Pilz2022}
Pilz, K.S., Lou, H.: Contextual and own-age effects in age perception.
  Scientific Reports  \textbf{12}(1),  14156 (2022).
  \doi{10.1038/s41598-022-18577-8}

\bibitem{ribeiro2016why}
Ribeiro, M.T., Singh, S., Guestrin, C.: " why should i trust you?" explaining
  the predictions of any classifier. In: Proceedings of the 22nd ACM SIGKDD
  international conference on knowledge discovery and data mining. pp.
  1135--1144 (2016)

\bibitem{rothe2015}
Rothe, R., Timofte, R., Van~Gool, L.: {DEX}: {Deep EXpectation} of apparent age
  from a single image. In: IEEE International Conference on Computer Vision
  Workshops (ICCVW). pp. 10--15 (2015)

\bibitem{terhorst2019reliable}
Terh{\"o}rst, P., Huber, M., Kolf, J.N., Zelch, I., Damer, N., Kirchbuchner,
  F., Kuijper, A.: Reliable age and gender estimation from face images: Stating
  the confidence of model predictions. In: 2019 IEEE 10th International
  Conference on Biometrics Theory, Applications and Systems (BTAS). pp.~1--8.
  IEEE (2019)

\bibitem{wen2018flipout}
Wen, Y., Vicol, P., Ba, J., Tran, D., Grosse, R.: Flipout: Efficient
  pseudo-independent weight perturbations on mini-batches. In: International
  Conference on Learning Representations (2018)

\bibitem{wimmer2023quantifying}
Wimmer, L., Sale, Y., Hofman, P., Bischl, B., H{\"u}llermeier, E.: Quantifying
  aleatoric and epistemic uncertainty in machine learning: Are conditional
  entropy and mutual information appropriate measures? In: Uncertainty in
  Artificial Intelligence. pp. 2282--2292. PMLR (2023)

\bibitem{zhifei2017cvpr}
Zhang, Z., Song, Y., Qi, H.: Age progression/regression by conditional
  adversarial autoencoder. In: IEEE Conference on Computer Vision and Pattern
  Recognition (CVPR). IEEE (2017)

\end{thebibliography}

\clearpage
\appendix

\section{Ethical Considerations}
Facial age estimation can be a controversial topic, as age has historically been used to discriminate persons and age can be private information. The aim of this research is to explore uncertainty estimation in a task that itself is inherently uncertain, humans are not very good at estimating a person's age from just a photo, requiring the concept of apparent age. Machine Learning models can to some degree estimate a person's age from a facial image, but models are not better than humans and should not be used for any kind of automated processing or discriminatory activities.

Uncertainty estimation usually improves the trustworthiness of a model, as the model can now express a confidence interval for the answer instead of a point estimate, and a human user receives information about the model being not sure about its output, and this information can be used to weight the usefulness of a prediction.

This paper only uses data from the APPA-REAL dataset and no human subjects were directly involved in this research.

\section{Additional Results and Details}

\begin{table*}[!ht]
\caption{Aleatoric and Epistemic Uncertainty by Model and Training  Data Percentage}
\label{tab:combined_ae}
\centering
\begin{tabular}{rcccccc}
\toprule
\textbf{Train Data Per. (\%)} & \multicolumn{2}{c}{\textbf{DROPCONNECT}} & \multicolumn{2}{c}{\textbf{ENSEMBLE}} & \multicolumn{2}{c}{\textbf{FLIPOUT}} \\
 & Aleatoric & Epistemic & Aleatoric & Epistemic & Aleatoric & Epistemic \\
\midrule
1   & 1.340  & 12.251 & 2.018  & 1.923  & 8.364  & 2.433  \\
5   & 19.855 & 12.891 & 13.902 & 1.862  & 2.404  & 7.072  \\
10  & 18.668 & 12.019 & 18.741 & 1.815  & 18.434 & 8.725  \\
25  & 18.286 & 10.890 & 18.601 & 1.783  & 17.655 & 8.994  \\
50  & 19.243 & 7.472  & 18.707 & 1.655  & 18.375 & 8.764  \\
75  & 18.865 & 6.006  & 18.890 & 1.600  & 18.499 & 8.536  \\
100 & 18.838 & 6.054  & 18.870 & 1.498  & 18.144 & 8.176  \\
\bottomrule
\end{tabular}
\end{table*}

\end{document}